\documentclass[cameraready]{Interspeech}

\title{MEG-to-MEG Transfer Learning and Cross-Task Speech/Silence Detection with Limited Data}

\author[affiliation={1}, orcid=0009-0001-9309-8507, correspondingauthor]{Xabier}{de Zuazo}
\author[affiliation={2}, orcid=0009-0001-3018-4731]{Vincenzo}{Verbeni}
\author[affiliation={1}, orcid=0000-0003-3804-4984]{Eva}{Navas}
\author[affiliation={1}, orcid=0000-0001-7282-2765]{Ibon}{Saratxaga}
\author[affiliation={2,4,5,6}, orcid=0000-0003-1694-5087]{Mathieu}{Bourguignon}
\author[affiliation={2,3}, orcid=0000-0002-7549-6042]{Nicola}{Molinaro}

\address{
    $^1$ HiTZ Center, University of the Basque Country -- UPV/EHU, Spain \\
    $^2$ Basque Center on Cognition, Brain and Language -- BCBL, Spain \\
    $^3$ Ikerbasque, Basque Foundation for Science, Spain \\
    $^4$ Laboratory of Functional Anatomy, Faculty of Human Motor Sciences, Université libre de Bruxelles (ULB), Belgium \\
    $^5$ Laboratoire de Neuroanatomie et Neuroimagerie translationnelles (LN2T), ULB Neuroscience Institute, Université libre de Bruxelles (ULB), Belgium \\
    $^6$ WEL Research Institute, Belgium
}

\email{xabier.dezuazo@ehu.eus, v.verbeni@bcbl.eu, eva.navas@ehu.eus, ibon.saratxaga@ehu.eus, mabourgu@ulb.ac.be, n.molinaro@bcbl.eu}

\keywords{transfer learning, MEG, speech decoding, cross-task decoding, brain-computer interface, neuroscience}

\usepackage{comment}

\begin{document}

\maketitle

\begin{abstract}
    Data-efficient neural decoding is a central challenge for speech brain-computer interfaces. We present the first demonstration of transfer learning and cross-task decoding for MEG-based speech models spanning perception and production. We pre-train a Conformer-based model on 50 hours of single-subject listening data and fine-tune on just 5 minutes per subject across 18 participants. Transfer learning yields consistent improvements, with in-task accuracy gains of 1-4\% and larger cross-task gains of up to 5-6\%. Not only does pre-training improve performance within each task, but it also enables reliable cross-task decoding between perception and production. Critically, models trained on speech production decode passive listening above chance, confirming that learned representations reflect shared neural processes rather than task-specific motor activity.
\end{abstract}

\section{Introduction}

Brain-computer interfaces (BCIs) for speech restoration require robust neural decoders~\cite{king2020, dezuazo2024_2}, but collecting sufficient training data per subject remains a fundamental challenge~\cite{anumanchipalli2019, moses2019, silva2024}. Nevertheless, while large-scale datasets have enabled substantial progress in non-invasive speech decoding tasks (including speech detection and phoneme decoding) using magnetoencephalography (MEG)~\cite{elvers2025, ozdogan2025, landau2025} and electroencephalography (EEG)~\cite{sato2024, dascoli2024}, practical BCI deployment is constrained by the limited data available per individual: typically minutes rather than hours. Current approaches train separate models from scratch for each subject and task~\cite{dash2018_1, dash2018_2, dezuazo2024_1, dezuazo2025_2}, ignoring the plausible benefits of transfer learning that have revolutionized computer vision~\cite{lecun2015} and natural language processing~\cite{vaswani2017}. Notwithstanding successful demonstrations of transfer learning in EEG~\cite{banville2025} and fMRI~\cite{benchetrit2024}, and recent advances in self-supervised neural decoding~\cite{defossez2023, jayalath2025}, transfer learning has only very recently been explored for MEG imagined speech using ImageNet-pretrained vision models~\cite{jhilal2026}, and remains untested for MEG-to-MEG pre-training and cross-task transfer between perception and production.

This study presents the first successful application of transfer learning to a MEG speech decoding task (speech detection). We pre-train a Conformer-based model~\cite{gulati2020, dezuazo2025_1} on 50 hours of single-subject listening data (LibriBrain~\cite{ozdogan2025}) and fine-tune on just 5 minutes per subject across 18 participants performing speech perception (listening and playback) and speech production tasks~\cite{bourguignon2020}. To our knowledge, we also provide the first cross-task MEG speech detection results, extending prior cross-task decoding work in non-speech EEG/MEG~\cite{magnabosco2024, aristimunha2025}. We show that transfer learning improves both in-task performance and cross-task generalization, with gains of 1-6\% across metrics. Importantly, models trained on production successfully decode passive listening above chance, confirming that decoding relies on shared neural speech representations rather than task-specific motor activity alone~\cite{philip2022, philip2023}. Together, these findings suggest that transfer learning may support more data-efficient MEG speech detection and improved cross-task generalization.

\section{Related Work}

MEG-based speech decoding has progressed from closed-vocabulary word classification~\cite{dash2018_1, dash2018_2, dash2019} to phone-level analysis~\cite{dezuazo2024_1, dezuazo2025_2, suzuki2025} and recent large-scale competitions~\cite{landau2025, elvers2025}. Complementary work has explored Transformer-based encoding models~\cite{klimovichgray2023}, compact end-to-end architectures~\cite{sarma2024}, and speech synthesis from MEG~\cite{kwon2024}. In parallel, EEG research has demonstrated open-vocabulary decoding~\cite{sato2024, accou2023, xu2024} and large-scale word and phonetic decoding~\cite{dascoli2024}, emphasizing the importance of dataset size and model capacity.

Transfer learning has been successfully applied to other neuroimaging modalities. EEG studies have leveraged self-supervised pre-training~\cite{banville2025} and contrastive learning~\cite{defossez2023} to improve cross-subject generalization. Recent work has combined discriminative decoders with language-model rescoring~\cite{jayalath2025} and foundation-model guidance~\cite{benchetrit2024} to improve performance. In contrast, despite these advances in EEG and fMRI, transfer learning has not been demonstrated for MEG-based speech decoding, leaving a critical gap in understanding whether large-scale pre-training can address the limited per-subject data available in clinical BCI settings.

Cross-task speech decoding between perception and production remains underexplored. At the same time, previous work established differences in neural representations between passive listening and overt speech~\cite{bourguignon2020, schoffelen2019, gwilliams2023}, and investigation of cross-task transfer and the role of motor-related activity versus shared speech representations has been limited~\cite{philip2023, levy2025}. Our work addresses this gap by demonstrating bidirectional cross-task decoding and quantifying the benefits of transfer learning across speech modalities.

\section{Methods}
\subsection{Datasets}

We evaluate transfer learning across two MEG speech datasets with contrasting characteristics: a large-scale single-subject dataset for pre-training and a multi-subject dataset with limited per-subject data for fine-tuning and cross-task evaluation.

\textbf{LibriBrain (pre-training).} We pre-train a single model on the LibriBrain dataset~\cite{ozdogan2025}, which provides over 50 hours of within-subject MEG recordings from a single participant during naturalistic English speech listening (Sherlock Holmes audiobooks). Recordings were acquired from a single right-handed male native English speaker using a 306-channel Elekta/MEGIN system (102 magnetometers, 204 planar gradiometers) and processed following the LibriBrain Competition pipeline~\cite{landau2025}, with model inputs downsampled to \SI{250}{Hz}. We use the binary Speech Detection task, which distinguishes speech from silence based on voice-activity labels. Across the dataset, speech accounts for approximately 76.7\% of the labeled time.

\textbf{Speech listening and production data~\cite{bourguignon2020} (fine-tuning and evaluation).} We fine-tune and evaluate on the dataset described by Bourguignon et al., comprising 18 healthy adult participants (9 female, 8 male, 1 unreported; mean age: 23.9 years) who were native Spanish speakers. Each participant performed three speech-related tasks for approximately 5 minutes each: listening to pre-recorded speech, listening to playback of their own voice, and reading aloud (speech production). MEG was recorded using another 306-channel Elekta/MEGIN system (same model as LibriBrain, but in a different lab), downsampled to \SI{250}{Hz} for consistency with LibriBrain. We formulate a similar speech detection task using voice-activity annotations aligned with the MEG signal. Across subjects, speech occupied on average \(78.6 \pm 2.3\%\) of frames during listening and \(74.8 \pm 3.9\%\) during playback (small differences reflect minor variations in usable recording durations), and \(75.6 \pm 3.7\%\) during speech production, with the remaining frames corresponding to silence. This dataset is not publicly available due to ethical constraints. Full experimental details are provided in~\cite{bourguignon2020}.

Crucially, the key distinction between datasets (50 hours of single-subject data versus 5 minutes per subject across 18 participants) enables us to investigate whether large-scale pre-training on one subject can improve decoding performance when fine-tuned on limited data from new subjects performing different speech tasks.

\subsection{Model Architecture}

We pretrained the MEGConformer~\cite{dezuazo2025_1}, a compact Conformer-based encoder~\cite{gulati2020} adapted for MEG time-series that operates directly on windowed raw sensor segments (306 channels) after preprocessing and downsampling to \SI{250}{Hz}. Accordingly, to match the limited amount of data available per subject and task in Bourguignon2020, we use \SI{0.5}{s} windows throughout and disable output smoothing, while keeping the remaining architectural choices and training recipe as close as possible to~\cite{dezuazo2025_1}.

For fine-tuning, we retain the original optimizer and most hyperparameters, but introduce three lightweight task-specific modifications:
(i) we select checkpoints using validation loss rather than F1-macro to avoid overfitting to a specific metric;
(ii) we introduce \textbf{RollAugment}, a fast roll-based temporal augmentation that circularly shifts each training frame by fixed fractions of the window (25\%, 50\%, and 75\%) and concatenates the shifted copies;
and (iii) we use soft targets given by the fraction of speech within each window (instead of hard 0/1 labels) to smooth the labels slightly.
Moreover, we use decimation by 4 to obtain \SI{250}{Hz} (via anti-aliased resampling) and reduce early-stopping patience to 10 epochs for efficiency.

\subsection{Experimental Protocol}

Subsequently, we pre-trained MEGConformer on LibriBrain's 50-hour listening dataset, then fine-tuned on Bourguignon2020 using two experimental paradigms:
(i) \textit{in-task}, where models were trained and evaluated on the same task (Listen, Playback, Production), and
(ii) \textit{cross-task}, where models were trained on one task and evaluated on a different task without retraining, covering all six possible train-test task pairings among Listen, Playback, and Production.
For each subject and condition, we compared transfer learning (pre-trained then fine-tuned) against training from scratch (no pre-training). Performance was assessed using F1-macro, balanced accuracy, and AUC-macro.

For LibriBrain pre-training, we used the official splits provided with the dataset~\cite{landau2025, ozdogan2025}. For Bourguignon2020, given the short duration of each recording and the absence of natural session boundaries, data were split independently for each subject and task into training (70\%), validation (15\%), and test (15\%) sets using random shuffling at the frame level. Accordingly, to normalize the signal, input windows were z-scored for each subject and task (separately for each sensor and time point). Training and validation data used training-set statistics, while test data were normalized using their own statistics to support fair in-task and cross-task evaluation. All training runs were performed on individual NVIDIA H100 GPUs.

Statistical significance was evaluated using the Wilcoxon signed-rank test~\cite{Wilcoxon1945}, a non-parametric method for paired comparisons when normality cannot be assumed~\cite{Santafe2015}. For each comparison, we tested whether the median improvement (transfer learning minus baseline) differed from zero (\(p~<~0.05\)), using subjects as the unit of replication. All p-values were corrected for multiple comparisons using the Holm-Bonferroni method~\cite{holm1979}. Overall transfer learning effects across metrics and task conditions were assessed with subject-level omnibus tests based on aggregated improvements, using a permutation-based sign-flip test (10,000 iterations)~\cite{pitman1937, good2005}.

\ifcameraready
    \textbf{Code availability.} To ensure reproducibility, all code, preprocessing scripts, and model configurations are publicly available at \url{https://github.com/hitz-zentroa/meg-phone-decoding}.
\else
    \textbf{Code availability.} To ensure reproducibility, all code, preprocessing scripts, and model configurations will be made publicly available; to preserve author anonymity, the link will be added after the review process.
\fi

\section{Results}
\subsection{In-Task Transfer Learning}

\begin{table}[th]
  \caption{In-task evaluation results (mean \(\pm\) std across subjects, in \%). Improvements over baseline are marked in bold.}
  \label{tab:intask_results}
  \centering
  \begin{tabular}{l l r r r}
    \toprule
    \multicolumn{2}{l}{\textbf{Task}} & \multicolumn{1}{c}{\textbf{Accu. (\%)}} & \multicolumn{1}{c}{\textbf{F1 (\%)}} & \multicolumn{1}{c}{\textbf{AUC (\%)}} \\
    \midrule
    \multicolumn{5}{l}{\textbf{Scratch (baseline)}} \\
     & \textit{listen} & \textit{76.2 \(\pm\) 4.8} & \textit{85.5 \(\pm\) 3.2} & \textit{64.0 \(\pm\) 9.4} \\
     & \textit{playback} & \textit{75.1 \(\pm\) 6.1} & \textit{84.0 \(\pm\) 4.7} & \textit{67.7 \(\pm\) 4.2} \\
     & \textit{production} & \textit{83.6 \(\pm\) 5.4} & \textit{89.7 \(\pm\) 3.5} & \textit{81.1 \(\pm\) 8.6} \\
    \addlinespace
    \multicolumn{5}{l}{\textbf{Transfer Learning}} \\
     & listen & \textbf{79.0 \(\pm\) 4.8} & \textbf{87.7 \(\pm\) 3.2} & \textbf{68.7 \(\pm\) 6.2} \\
     & playback & \textbf{76.0 \(\pm\) 6.0} & \textbf{85.4 \(\pm\) 4.4} & 67.2 \(\pm\) 8.6 \\
     & production & \textbf{84.2 \(\pm\) 4.9} & \textbf{90.3 \(\pm\) 3.4} & \textbf{82.0 \(\pm\) 8.0} \\
    \addlinespace
    \bottomrule
  \end{tabular}
\end{table}

Table~\ref{tab:intask_results} presents in-task performance comparing models trained from scratch on Bourguignon2020 against models pre-trained on LibriBrain and fine-tuned on the same data. Transfer learning improved performance across nearly all metrics and tasks. For the listening task, transfer learning yielded significant gains of +3.7\% accuracy, +2.6\% F1, and +7.3\% AUC (\(W=17.0\), \(p=0.005\)). The playback task showed more modest, non-significant improvements of +1.2\% accuracy, +1.7\% F1, and -0.7\% AUC (\(W=45.0\), \(p=0.163\)). Interestingly, despite being pre-trained exclusively on a listening task, transfer learning improved all metrics for speech production, including +0.7\% accuracy, +0.7\% F1, and +1.1\% AUC, though these differences did not reach statistical significance (\(W=61.0\), \(p=0.304\)). Across all tasks and metrics, a sign-flip permutation test confirmed a significant overall difference of transfer learning (\(p~<~0.001\)). These results demonstrate that representations learned from large-scale single-subject listening data generalize to new subjects and tasks, even with only ~5 minutes of data per subject.

\subsection{Cross-Task Decoding Baseline}

\begin{table}[t]
  \caption{Cross-task decoding results for the model trained from scratch (mean \(\pm\) std across subjects, in \%). All reported cross-task results are significantly above chance.}

  \label{tab:xtask_scratch}
  \centering
  \begin{tabular}{l r r r}
    \toprule
    \textbf{Train \(\rightarrow\) Test} & \multicolumn{1}{c}{\textbf{Accu. (\%)}} & \multicolumn{1}{c}{\textbf{F1 (\%)}} & \multicolumn{1}{c}{\textbf{AUC (\%)}} \\
    \midrule
    listen to play. & 72.5 \(\pm\) 3.6 & 83.0 \(\pm\) 2.7 & 59.4 \(\pm\) 5.3 \\
    listen to prod. & 71.5 \(\pm\) 3.5 & 82.3 \(\pm\) 2.6 & 57.9 \(\pm\) 7.4 \\
    play. to listen & 73.4 \(\pm\) 2.9 & 83.7 \(\pm\) 2.2 & 59.4 \(\pm\) 5.7 \\
    play. to prod. & 71.4 \(\pm\) 4.4 & 82.1 \(\pm\) 3.6 & 57.4 \(\pm\) 6.7 \\
    prod. to listen & 66.1 \(\pm\) 6.3 & 77.7 \(\pm\) 5.8 & 54.0 \(\pm\) 6.3 \\
    prod. to play. & 65.0 \(\pm\) 5.9 & 76.5 \(\pm\) 5.5 & 54.6 \(\pm\) 4.7 \\
    \bottomrule
  \end{tabular}
\end{table}

To establish whether speech representations transfer across tasks, we evaluated models trained on one task and tested on another without transfer learning. Table~\ref{tab:xtask_scratch} presents all six cross-task train-test pairs for models trained from scratch on Bourguignon2020. All cross-task decodings were individually significant across all metrics (paired Wilcoxon signed-rank test against chance, all \(p~<~0.05\)), and overall cross-task performance was highly significant according to a permutation test (\(p~<~0.001\)). This demonstrates that neural representations for speech processing are partially shared across listening, playback, and production tasks. Cross-task accuracy ranged from 65.0\% to 73.4\%, with perception tasks (listening and playback) transferring more effectively to each other (72.5-73.4\%) than production transferring to perception tasks (65.0-66.1\%).

\subsection{Cross-Task with Transfer Learning}

\begin{table}[th]
  \caption{Cross-task decoding results for the pretrained model (mean \(\pm\) std across subjects, in \%). Improvements over baseline are marked in bold.}
  \label{tab:xtask_pretrained}
  \centering
  \begin{tabular}{l r r r}
    \toprule
    \textbf{Train \(\rightarrow\) Test} & \multicolumn{1}{c}{\textbf{Accu. (\%)}} & \multicolumn{1}{c}{\textbf{F1 (\%)}} & \multicolumn{1}{c}{\textbf{AUC (\%)}} \\
    \midrule
    listen to play. & \textbf{76.9 \(\pm\) 4.0} & \textbf{86.5 \(\pm\) 2.8} & \textbf{61.1 \(\pm\) 4.5} \\
    listen to prod. & \textbf{75.3 \(\pm\) 5.2} & \textbf{85.3 \(\pm\) 3.9} & 57.3 \(\pm\) 7.3 \\
    play. to listen & \textbf{78.0 \(\pm\) 3.9} & \textbf{87.1 \(\pm\) 2.9} & \textbf{61.6 \(\pm\) 5.8} \\
    play. to prod. & \textbf{73.6 \(\pm\) 7.1} & \textbf{83.7 \(\pm\) 6.1} & \textbf{58.1 \(\pm\) 9.1} \\
    prod. to listen & \textbf{69.3 \(\pm\) 8.0} & \textbf{80.1 \(\pm\) 7.3} & \textbf{56.0 \(\pm\) 5.7} \\
    prod. to play. & \textbf{68.3 \(\pm\) 7.4} & \textbf{79.0 \(\pm\) 7.2} & \textbf{56.6 \(\pm\) 3.8} \\
    \bottomrule
  \end{tabular}
\end{table}

Table~\ref{tab:xtask_pretrained} presents cross-task decoding performance with transfer learning from LibriBrain. Pre-training substantially improved cross-task generalization across nearly all task combinations and metrics. As expected, perception tasks showed the strongest gains: listen-to-playback improved by +6.1\% accuracy, +4.2\% F1, and +2.9\% AUC (\(W=3.0\), \(p~<~0.001\)), while playback-to-listen improved by +6.3\% accuracy, +4.1\% F1, and +3.7\% AUC (\(W=3.0\), \(p~<~0.001\)). Cross-task decoding involving production also benefited significantly from transfer learning. Listen-to-production improved by +5.3\% accuracy and +3.6\% F1 (\(W=22.0\), \(p=0.016\)), while production-to-listen gained +4.8\% accuracy, +3.1\% F1, and +3.7\% AUC (\(W=36.0\), \(p=0.048\)). Similarly, production-to-playback improved by +5.1\% accuracy, +3.3\% F1, and +3.7\% AUC (\(W=33.0\), \(p=0.048\)). Across all cross-task combinations and metrics, a sign-flip permutation test confirmed a highly significant overall benefit of transfer learning (\(p~<~0.001\)). These results show that large-scale pre-training improves cross-task speech detection even with minimal per-subject fine-tuning data, and that the improvements extend beyond perception-only tasks to include speech production.

\begin{figure}[t]
  \centering
  \includegraphics[width=\linewidth]{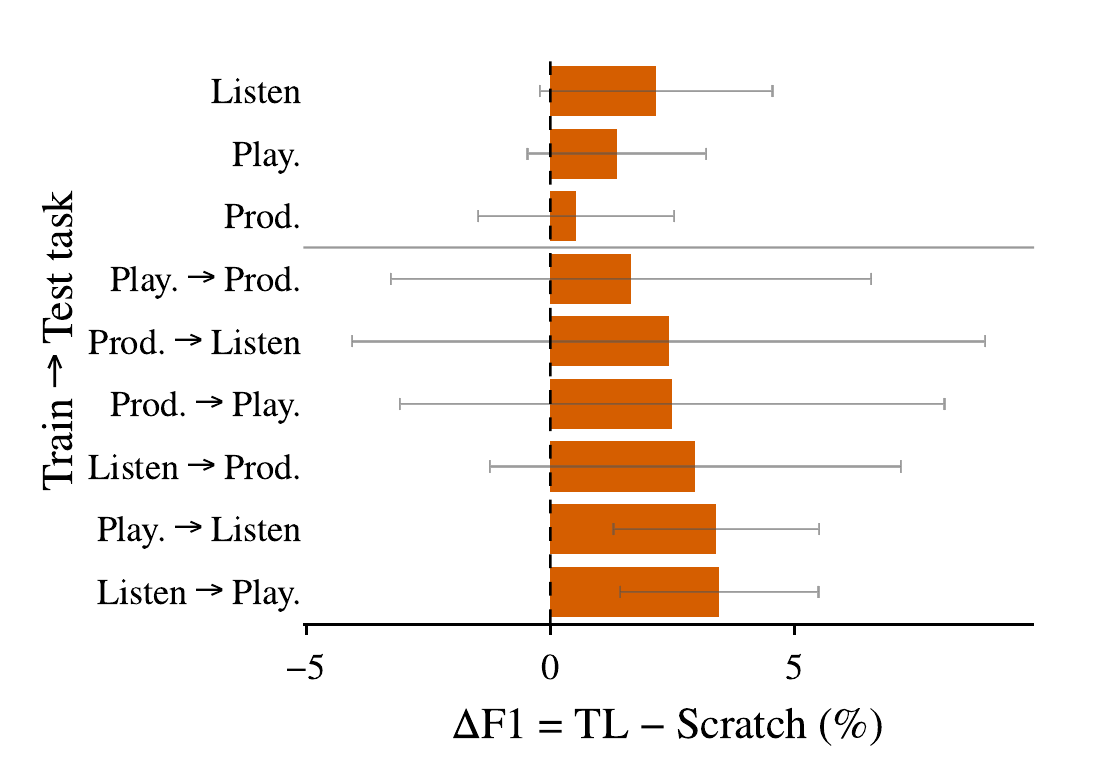}
  \caption{Effect size of transfer learning for cross-task decoding. Bars show the mean improvement in F1 score (transfer learning minus training from scratch) for each task and train-test task pair. Error bars indicate the standard deviation across subjects.}
  \label{fig:xtask_effect_size_f1}
\end{figure}

\begin{figure}[t]
  \centering
  \includegraphics[width=\linewidth]{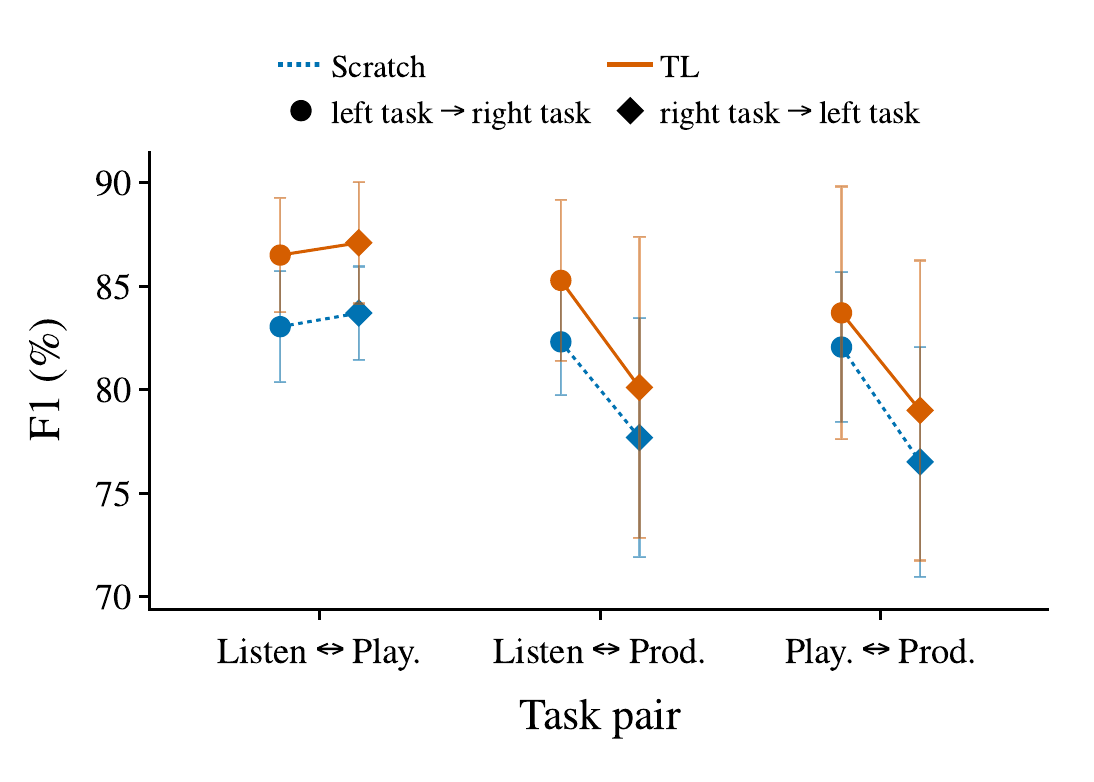}
  \caption{Cross-task F1 asymmetry for each task pair. Each x-axis category shows two directions: circle markers correspond to the left-to-right direction in the label, and diamond markers to the reverse direction. For each model, the two direction-specific points are connected to highlight asymmetry. Error bars indicate the standard deviation across subjects.}
  \label{fig:xtask_asymmetry_f1}
\end{figure}

\begin{figure}[t]
  \centering
  \includegraphics[width=\linewidth]{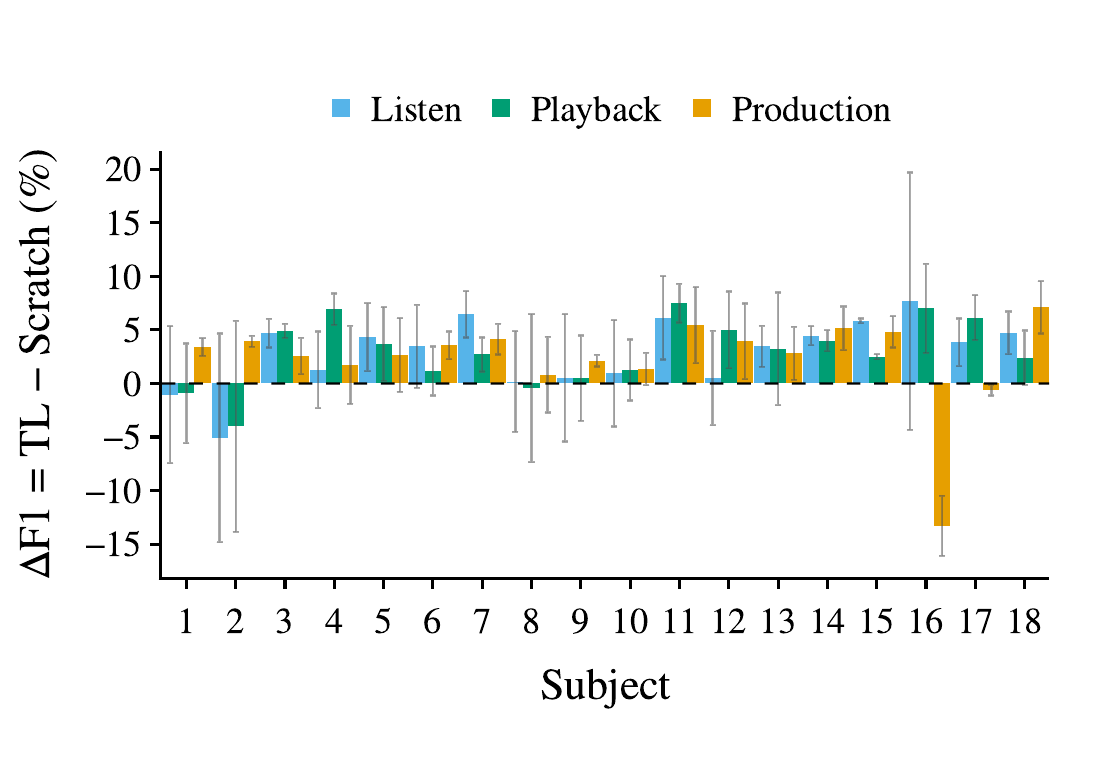}
  \caption{Subject-level in-task F1 improvement with transfer learning. Each bar shows the improvement in F1 score (transfer learning minus training from scratch) for a single subject and task. Colors indicate the three tasks.}
  \label{fig:xtask_f1_per_subject}
\end{figure}

Figure~\ref{fig:xtask_effect_size_f1} summarizes the effect size of transfer learning across all conditions. Notably, improvements were observed throughout, with more modest gains for in-task decoding (0.5-2.2\%) and larger gains for cross-task decoding (1.7-3.5\%), and with cross-task decoding involving production showing greater variability (error bars). Figure~\ref{fig:xtask_asymmetry_f1} reveals systematic asymmetries in bidirectional cross-task transfer. While listening and playback transferred bidirectionally with similar performance (Listen to Playback: 86.5\% F1, Playback to Listen: 87.1\% F1), cross-task combinations involving production showed clear directional preferences. Perception-to-production transfer (Listen to Production: 85.3\% F1, Playback to Production: 83.7\% F1) substantially outperformed production-to-perception transfer (Production to Listen: 80.1\% F1, Production to Playback: 79.0\% F1). This asymmetry likely reflects the fact that production naturally involves auditory self-monitoring, whereas perception tasks do not engage motor planning representations~\cite{hickok2007, houde2011}. However, the above-chance performance of production-to-perception decoding confirms that production models capture shared neural speech representations beyond production-specific representations. The subject-level analysis in Figure~\ref{fig:xtask_f1_per_subject} shows predominantly positive transfer effects, though with substantial individual variability. In perception tasks, 15 of 18 subjects showed improvements, while 2 (1, 2) exhibited clear negative effects. For production, 16 of 18 subjects benefited from transfer learning, though Subject 16 showed a marked negative effect (-13.3\% F1). Consequently, this inter-subject variability suggests that transfer learning effectiveness may depend on individual neural organization or recording quality, highlighting the need for subject-adaptive approaches in practical BCI applications.

\section{Discussion and Conclusion}

In this work, we demonstrate that MEG-to-MEG transfer learning from large-scale single-subject MEG data enhances a MEG speech decoding task (speech detection) across new subjects and tasks, even with only a few minutes of subject-specific data. Pre-training on LibriBrain consistently improved in-task performance and, more importantly, substantially improved cross-task generalization between speech perception and production. This suggests that the MEGConformer learns representations that capture core neural processes underlying speech, rather than relying solely on task or subject-specific patterns.

Specifically, a key finding is the asymmetric nature of cross-task transfer. While perception tasks (listening and playback) transferred bidirectionally with comparable performance, decoding from production to perception was consistently weaker than the reverse direction. This asymmetry is expected, given that speech production engages additional processes--such as motor planning, efference copy, and somatosensory feedback--that are absent during passive perception, whereas perceptual representations form a core subset of those recruited during production. As a consequence, models trained on perception data can efficiently decode speech production tasks, whereas models trained on production must disentangle perceptual information from concurrent motor-related activity, yielding weaker transfer in the production-to-perception direction. Nonetheless, the ability of production-trained models to decode listening and playback tasks at above-chance levels provides further evidence that speech perception and production recruit overlapping neural circuitry, as indicated by interactions between ventral perceptual pathways and dorsal sensorimotor circuits that support contemporary dual-stream models of speech processing~\cite{hickok2004, pulvermuller2006, hickok2007, houde2011, cogan2014}.

From a practical perspective, our results highlight the feasibility of training effective MEG speech detection models with very limited per-subject data. Pre-training on LibriBrain consistently improved in-task performance, with larger gains observed in cross-task generalization between speech perception and production. In fact, statistically reliable gains emerged primarily in cross-task decoding rather than in-task performance. This is a critical step toward practical neurotechnology applications, where long calibration sessions are often infeasible.

Nevertheless, several limitations should be noted. First, we focus on a speech detection task, which does not capture higher-level phoneme, word, or semantic representations, and pre-training and fine-tuning involved different languages (English and Spanish). Second, pre-training relied on a single subject, and future work should assess whether multi-subject pre-training improves generalization. Finally, while improvements are consistent, their magnitude is modest and variable, indicating that transfer learning complements rather than replaces subject-specific adaptation. Looking ahead, future work will extend this approach to more complex speech tasks, including phone classification~\cite{dezuazo2024_1, dezuazo2025_2}, keyword spotting~\cite{elvers2025}, and speech synthesis~\cite{anumanchipalli2019}, and will explore transfer learning and cross-task decoding across larger and more diverse datasets.

\ifcameraready
    \section{Acknowledgements}

    This research was supported by the Basque and Spanish Governments (grant IKUR-IKA-23/18, and project AIA2025-163317-C31 funded by MICIU/AEI / 10.13039/501100011033/). We are grateful to the DIPC Supercomputing Center and EJIE, the technological management body of the Basque Government, for providing essential technical and human resources. This work was carried out within the \#neural2speech and brAIn2lang research teams.
\fi

\bibliographystyle{IEEEtran}
\bibliography{mybib}

\end{document}